\documentclass[10pt,twocolumn,letterpaper]{article}

\usepackage[pagenumbers]{cvpr} 

\usepackage[dvipsnames]{xcolor}

\usepackage{bbding}
\usepackage{multicol}
\usepackage{multirow}
\usepackage{makecell}

\definecolor{cvprblue}{rgb}{0.21,0.49,0.74}
\usepackage[pagebackref,breaklinks,colorlinks,citecolor=cvprblue]{hyperref}
\usepackage{graphicx}

\def\paperID{} 
\def\confName{3DV\xspace}
\def\confYear{2026\xspace}

\title{Learnable SMPLify: A Neural Solution for Optimization-Free Human Pose Inverse Kinematics}

\author{Yuchen Yang\thanks{Work performed during his internship at Shanghai Artificial Intelligence Laboratory.} \textsuperscript{~1,2}
\quad Linfeng Dong\textsuperscript{3,2}
\quad Wei Wang\textsuperscript{2} 
\quad Zhihang Zhong\textsuperscript{2}
\quad Xiao Sun\textsuperscript{\Envelope~2}\\
\textsuperscript{1}Fudan University \quad \textsuperscript{2}Shanghai Artificial Intelligence Laboratory \quad
\textsuperscript{3}Zhejiang University} %

\begin{document}
\maketitle
\begin{abstract}
In 3D human pose and shape estimation, SMPLify remains a robust baseline that solves inverse kinematics (IK) through iterative optimization. However, its high computational cost limits its practicality.
Recent advances across domains have shown that replacing iterative optimization with data-driven neural networks can achieve significant runtime improvements without sacrificing accuracy.
Motivated by this trend, we propose Learnable SMPLify, a neural framework that replaces the iterative fitting process in SMPLify with a single-pass regression model. The design of our framework targets two core challenges in neural IK: data construction and generalization. To enable effective training, we propose a temporal sampling strategy that constructs initialization–target pairs from sequential frames. To improve generalization across diverse motions and unseen poses, we propose a human-centric normalization scheme and residual learning to narrow the solution space.
Learnable SMPLify supports both sequential inference and plug-in post-processing to refine existing image-based estimators.
Extensive experiments demonstrate that our method establishes itself as a practical and simple baseline: it achieves nearly 200× faster runtime compared to SMPLify, generalizes well to unseen 3DPW and RICH, and operates in a model-agnostic manner when used as a plug-in tool on LucidAction.
The code is available at \url{https://github.com/Charrrrrlie/Learnable-SMPLify}.
\end{abstract}    
\section{Introduction}
\label{sec:intro}
The replacement of traditional iterative optimization with data-driven and learned inference via neural networks has become a unifying trend across many domains.
For example, VGGT~\cite{wang2025vggt} in multi-view stereo, GAIL~\cite{ho2016generative} in control, and even NeuroSAT~\cite{selsam2018learning} for satisfiability solving.
In 3D human pose and shape estimation, classical optimization-based approaches such as SMPLify(-X)~\cite{smplify,smplifyx} continue to serve as robust baselines and are still widely employed in downstream tasks such as 3D reconstruction and animation~\cite{qiu2025lhm,yu2023monohuman,niu2025anicrafter}.
This enduring reliance on SMPLify(-X) motivates the need for research to enhance its performance, running in parallel with image-based regression methods~\cite{kanazawa2018end,goel2023humans,cai2023smpler,baradel2024multi,wang2025prompthmr}.

\begin{figure}[t]
    \centering
    \includegraphics[width=\linewidth]{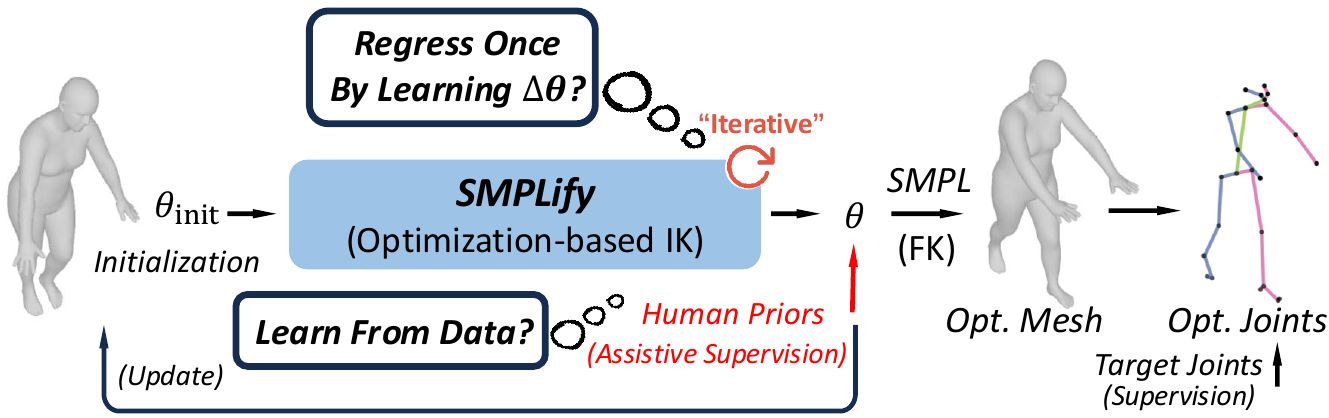}
    \caption{Illustration of SMPLify and motivation of our work. IK and FK denote inverse kinematics and forward kinematics, respectively. Opt. refers to optimized items. The high computational cost of the iterative process and the reliance on human priors motivate the development of a data-driven and non-iterative approach.}
    \label{fig:teaser}
\end{figure}

As illustrated in \cref{fig:teaser}, SMPLify addresses the ill-posed inverse kinematics (IK) problem of fitting the SMPL model to provided joints, aiming to minimize joint reconstruction error while enforcing human body priors.
Being fundamentally optimization-based, SMPLify suffers from two functional limitations.
Firstly, the performance heavily depends on initialization quality.
To improve this, existing methods leverage rich image cues by training neural networks to produce better initialization~\cite{kolotouros2019spin,guler2019holopose,joo2021exemplar}, and by incorporating various auxiliary constraints such as silhouettes and part segmentations~\cite{huang2017towards,lassner2017unite,rempe2021humor,xiang2019monocular}, thereby stabilizing optimization.
Additionally, SMPLify relies on time-consuming iterative procedures. In response, hybrid IK methods attempt to obtain a closed-form solution from images by estimating elements like twist angles~\cite{li2021hybrik,li2023niki} or vertex maps~\cite{shetty2023pliks}.

However, solving the core SMPLify inverse kinematics problem directly from joints, without auxiliary image cues, remains underexplored.
Song \etal~\cite{song2020human} employ gradient information to guide IK but still require iterative procedures. KAMA~\cite{iqbal2021kama} derives an analytical solution by ignoring twist rotation ambiguity, which compromises accuracy.
No existing method provides a solution that is both accurate and entirely free of optimization.

In this work, we propose Learnable SMPLify, a neural inverse kinematics framework that estimates SMPL parameters from joints through a single-time regression. Our framework aligns the overall setting of SMPLify but without iterative optimization, while maintaining high accuracy and anatomical validity by a data-driven way.

The detailed design of Learnable SMPLify specifically targets two core challenges in neural inverse kinematics.
1) Proper initialization-target pair data for training. Initialization requires being sufficiently close to the target to ensure convergence in the ill-posed IK space, yet not too close to avoid trivial learning.
To balance this, we exploit the spatial continuity of human motion and construct initialization-target pairs from adjacent video frames. By varying the temporal interval between them, the model learns to transfer initial SMPL parameters toward the target across a spectrum of realistic and plausible data.
2) Generalization ability. Unlike optimization-based IK solvers that adapt to each instance through iterative fitting, neural IK models must generalize across diverse motions and unseen poses. 
To stabilize inference, we design a human-centric coordinate system to normalize initial and target joints. This normalization eliminates variations in global orientation and translation.
Furthermore, instead of directly regressing absolute SMPL parameters, the network predicts the residuals between initialization and target, which constrains the solution space and further enhances generalization.
To sum up, Learnable SMPLify collects initial and target joints from the adjacent video sequences and normalizes them in the human-centric coordinate system. The normalized joints are passed to a GCN-based joint feature extractor~\cite{yan2018spatial}, which encodes motion-aware features. These features are then used to regress the residual SMPL parameters between the initialization and the target, producing the final results in a single forward pass.

Trained on AMASS~\cite{mahmood2019amass} and evaluated on large-scale AMASS, 3DPW~\cite{von2018pw3d}, and RICH~\cite{huang2022rich} datasets, the proposed Learnable SMPLify framework consistently outperforms existing methods. Compared to the classical SMPLify, Learnable SMPLify is nearly $\times200$ times faster and achieves over $5mm$ PVE improvement across all datasets.
The results validate the effectiveness of our approach and are coherent with the accurate and optimization-free motivation.
Furthermore, we demonstrate the utility of Learnable SMPLify as a plug-in post-processing module by refining predictions from existing image-based human pose and shape estimation methods~\cite{shen2024world,yin2025smplest} on the challenging LucidAction~\cite{dong2024lucidaction} dataset.

Our contributions are summarized into threefold:
\begin{itemize}
    \item We propose Learnable SMPLify, a neural inverse kinematics framework that regresses SMPL parameters from joints in a single forward pass, removing the need for iterative optimization while maintaining high accuracy.
    \item We introduce a human-centric normalization scheme and a training strategy based on temporally adjacent frames, enabling effective residual learning in the ill-posed inverse kinematics setting.
    \item We demonstrate that Learnable SMPLify serves as a simple yet strong baseline across multiple datasets, highlighting its practicality and generalization capability for future research in neural inverse kinematics.
\end{itemize}

\section{Related Work}
\label{sec:related-work}

\subsection{General Inverse Kinematics}
The inverse kinematics (IK) problem has been extensively studied in robotics, computer graphics, and related fields.
Classical numerical solvers~\cite{balestrino1984robust,klein2012review,buss2005selectively,wolovich1984computational} address the IK problem via iterative optimization, which is conceptually straightforward but often computationally expensive.
To improve efficiency, heuristic methods~\cite{luenberger1984linear,aristidou2011fabrik,rokbani2014ik} offer faster solutions by involving local adjustment and approximation, rather than global optimization.
More recently, data-driven approaches have gained popularity, where neural networks are trained to directly predict inverse solutions~\cite{csiszar2017mvip,zhou2019continuity,ardizzone2018analyzing}. 
Several neural network-based methods are tailored to introduce domain knowledge for particular tasks, such as human motion~\cite{villegas2018neural}, and pose editing~\cite{voleti2022smplik,oreshkin2021protores,jiang2024manikin}.
The task-specific nature highlights the effectiveness of vertical domain methods, motivating the exploration of learnable inverse kinematics formulations for human pose and shape estimation.

\subsection{Human Pose and Shape Inverse Kinematics}
\label{sec:human-ik}
We review human pose and shape (HPS) estimation methods from the perspective of IK, with a focus on approaches that utilize parametric models of SMPL(-X).
 
Traditionally, IK refers to computing joint angles given the desired position or orientation of specific body parts. 
In HPS, inverse kinematics extends to regression-based methods that predict SMPL parameters from images, where inferring pose and shape from observations inverts the SMPL forward process.
A large number of methods~\cite{kanazawa2018end,kocabas2020vibe,kocabas2021pare,kolotouros2019convolutional,guler2019holopose} fall into this category. To handle the diversity of in-the-wild images, recent approaches adopt larger neural networks trained on increasingly diverse datasets~\cite{goel2023humans,cai2023smpler,baradel2024multi,lin2023one,wang2025prompthmr}.
Meanwhile, PyMAF~\cite{zhang2021pymaf} and ReFit~\cite{wang2023refit} introduce iterative feedback mechanisms to refine predictions, incorporating the strengths of optimization-based HPS methods.

Optimization-based methods follow the conventional IK paradigm. With the proposal of SMPL(-X) model, SMPLify(-X)~\cite{smplify,smplifyx} formulates HPS as an optimization problem that iteratively fits SMPL(-X) model to the given keypoints.
The objective function typically combines joint fitting error with regularization terms based on pose and shape priors from kinematic rules and simple generative models~\cite{mclachlan2000finite, kingma2013auto}.
Although SMPLify(-X) serves as a robust baseline in practical applications, it is sensitive to initialization and suffers from slow convergence due to iterative optimization.
To stabilize optimization, existing methods leverage regression networks to produce better initialization~\cite{kolotouros2019spin,guler2019holopose,joo2021exemplar}, incorporate auxiliary constraints~\cite{huang2017towards,lassner2017unite,rempe2021humor,xiang2019monocular,tiwari2022pose}, and extend to complex application scenarios~\cite{zanfir2018monocular,patel2024camerahmr}.
To accelerate the iteration process, Li~\etal~\cite{li2021hybrik,li2023niki} estimate twist angles from images to enable a closed-form solution, while PLIKS~\cite{shetty2023pliks} introduces a pseudo-linear solver by predicting UV maps.

The setting of KAMA~\cite{iqbal2021kama} and Song~\etal~\cite{song2020human} is most closely related to our work. Both aim to accelerate the SMPLify IK process by operating directly on joint positions, without relying on additional visual input.
Song~\etal employ neural networks to incorporate the SMPL parameter gradient using a fixed number of parameter updates during the forward process. KAMA derives an analytical solution by applying approximation rules based on the kinematic tree.
However, Song~\etal still depend on the iterative process, while KAMA achieves efficiency at the cost of reduced accuracy.
In this paper, we accelerate the SMPLify optimization by replacing its iterative process with a single forward pass through a neural network. 
Our method eliminates the need for costly optimization while maintaining high accuracy, enabled by carefully designed initialization and normalization strategies.

\section{Method}
\subsection{Preliminaries}
\label{sec:preliminary}
The SMPL body model~\cite{loper2015smpl}, denoted as $\mathcal{M}$, parameterizes the 3D human mesh consisting of $N=6890$ vertices $V$ using pose parameters $\boldsymbol{\theta} \in \mathbb{R}^{24\times3}$ and shape parameters $\boldsymbol{\beta} \in \mathbb{R}^{10}$. The forward kinematics (FK) of SMPL is to deform the mesh template based on given pose and shape parameters:
\begin{equation}
    V = \mathcal{M}(\boldsymbol{\theta}, \boldsymbol{\beta}, \xi),
\label{eq:smpl-forward}
\end{equation}
where $\xi$ indicates the pose and shape blend parameters of SMPL model $\mathcal{M}$. 

SMPL is defined in a canonical space and can be transformed into world coordinates using translation $T$.
The joint locations $J$ can be computed from the mesh vertices $V$ using a fixed linear regressor $Jtr$:
\begin{equation}
    J = Jtr V.
\label{eq:smpl-regress}
\end{equation}
The conventional inverse kinematics (IK) of SMPL refers to estimating pose and shape parameters from joint positions with initialization:
\begin{equation}
    \boldsymbol{\theta}, \boldsymbol{\beta} = IK(J, \boldsymbol{\theta}_{init}, \boldsymbol{\beta}_{init}).
\label{eq:ik-definition}
\end{equation}
In the original SMPLify~\cite{smplify}, the initialization is set to T-pose, where both $\boldsymbol{\theta}_{init}$ and $\boldsymbol{\beta}_{init}$ are zero vectors.
However, due to information loss in the forward kinematics, the inverse kinematics problem is severely ill-posed when learning from a suboptimal initialization.
Prior studies~\cite{kolotouros2019spin,guler2019holopose,joo2021exemplar} demonstrate that improved initialization significantly enhances the performance of SMPLify.

\subsection{Learnable SMPLify}
\subsubsection{Problem Formulation}
Similar to prior optimization-based methods~\cite{li2021hybrik,li2023niki,iqbal2021kama}, we introduce Learnable SMPLify, an optimization-free neural solver that improves SMPLify with 3D joint locations $J_{target}$ as input.
As defined in \cref{eq:ik-definition}, our problem setup follows the conventional SMPL IK problem: the target joints $J_{target}$ and initial parameters $(\boldsymbol{\theta}_{init}, \boldsymbol{\beta}_{init})$ are given to predict SMPL parameters $(\boldsymbol{\theta}_{pred}, \boldsymbol{\beta}_{pred})$.
Since human shape parameters can be inferred from skeletal structure~\cite{iqbal2021kama}, we focus our efforts on regressing the pose parameters~\cite{li2021hybrik,li2023niki}. For simplicity, we assume that the initial and target shape parameters are considered identical:
\begin{equation}
\label{eq:assumption}
    \boldsymbol{\beta} \equiv \boldsymbol{\beta}_{init} \equiv \boldsymbol{\beta}_{target}.
\end{equation}
Its practical implementation is provided in \cref{sec:implementation}.

\begin{figure*}[t]
    \centering
    \includegraphics[width=\linewidth]{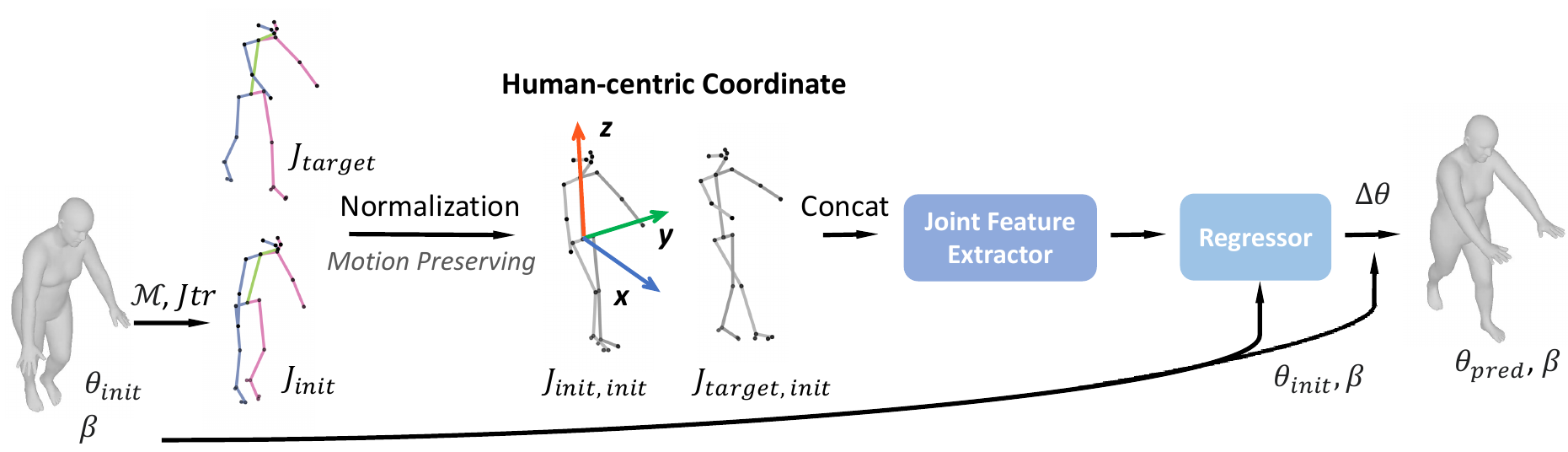}
    \caption{Overview of Learnable SMPLify. The framework first constructs a human-centric coordinate system from the initial joints and uses it to normalize both the initial and target joints into the same reference frame. Then, a neural solver, comprised of a joint feature extractor and a regressor, takes the normalized joints along with the initial SMPL parameters to predict residual pose parameters.}
    \label{fig:framework}
\end{figure*}

Formally, as illustrated in \cref{fig:framework}, Learnable SMPLify mimics the IK process through a learnable function $\mathcal{F}$:
\begin{equation}
    \boldsymbol{\theta}_{pred} = \mathcal{F}(J_{target}, \boldsymbol{\theta}_{init}, \boldsymbol{\beta}).
\end{equation}

\subsubsection{Framework}
\noindent
\textbf{Preprocessing.}
Given the initial parameters, we obtain
the initial joints $J_{init}$ via SMPL forward kinematic process, defined in \cref{eq:smpl-forward} and \cref{eq:smpl-regress}:
\begin{equation}
    J_{init} = Jtr \mathcal{M}(\boldsymbol{\theta}_{init}, \boldsymbol{\beta}, \xi).
\end{equation}

\vspace{1mm}
\noindent
\textbf{Mindset.}
With both the initial and target joint positions available, we reinterpret the IK problem as a motion transition task.
Specifically, regression from initial SMPL parameters to target ones can be guided by modeling the motion transition from initial joints to target joints.
Therefore,
Learnable SMPLify framework becomes straightforward: a feature extraction module that encodes information from joint positions and a regressor that predicts the residual between initial and target SMPL parameters.

\vspace{2mm}
\noindent
\textbf{Human-centric Coordinate.}
Unlike optimization-based IK solvers that perform fitting per instance, Learnable SMPLify must generalize across diverse joint inputs, varying in action, scale, and orientation.
To ease the data diversity issue, we propose a human-centric normalization strategy before joint feature extraction.

The normalization strategy constructs an orthonormal frame from joint locations, reducing the model's solution space by eliminating variations in global orientation and translation. Specifically, for a given skeleton of joints $J$, we define its human-centric coordinate system as follows:
First, we centralize the skeleton by translating all joints so that the pelvis lies at the origin. 
Let $J_p$ denote the pelvis joint, and define the translation:
\begin{equation}
    T = J_p,\ \ and \ \ J^c = J -T.
\end{equation}
Let $J_{lh}$, $J_{rh}$, $J_t$ denote left hip, right hip, and thorax of the centralized joints $J^c$, respectively.
We then define two direction vectors $\vec{t}_1$ and $\vec{t}_2$:
\begin{equation}
    \vec{t}_1 = ||\vec{J}_{lh}-\vec{J}_{rh}||,
\end{equation}
\begin{equation}
    \vec{t}_2 = ||\vec{J}_{p}-\vec{J}_{t}||.
\end{equation}
From these, we construct an orthonormal basis as follows.
The y-axis is defined as the unit vector along $\vec{t}_1$:
\begin{equation}
    \vec{y} = \frac{\vec{t_1}}{||\vec{t_1}||},
\end{equation}
The z-axis is the component of $\vec{t}_2$ orthogonal to $\vec{y}$ with normalization:
\begin{equation}
    \vec{z} = \frac{\vec{t}_2 - (\vec{t}_2 \cdot \vec{y}) \cdot \vec{y}}{||\vec{t}_2 - (\vec{t}_2 \cdot \vec{y}) \cdot \vec{y}||}.
\end{equation}
The x-axis is given by the normalized cross product to ensure a right-handed coordinate system:
\begin{equation}
    \vec{x} = \frac{\vec{z} \times \vec{y}}{||\vec{z} \times \vec{y}||}.
\end{equation}
Finally, we assemble the rotation matrix $R\in \mathbb{R}^{3\times3}$ that transforms vectors from the world coordinate system into this human-centric frame:
\begin{equation}
    R = \begin{bmatrix}
        \vec{x} \ \
        \vec{y} \ \
        \vec{z}
        \end{bmatrix}^T.
\end{equation}

With the human-centric coordinate frame defined above, we now apply it to transform joint positions for SMPL IK.
Using the rotation matrix $R_{init}$ and translation vector $T_{init}$ of the initialization joints $J_{init}$, we perform a rigid transformation, yielding the joint locations in human-centric coordinates of initialization, denoted as $J_{init, init}$.
For the target joints, we adopt the same coordinate system defined by $J_{init}$ and transform the target joints $J_{target}$ accordingly to obtain $J_{target, init}$.
Instead of converting $J_{target}$ to its own human-centeric coordinates, i.e. $J_{target, target}$, the current process maintains relative motion information from $J_{init}$ to $J_{target}$, facilitating meaningful feature extraction for target SMPL parameter regression.

\vspace{2mm}
\noindent
\textbf{Neural Solver.}
To regress target SMPL parameters, we employ a joint feature extractor and a regressor.
Note that we focus on the general formulation of neural IK frameworks. Therefore, we leverage existing well-structured networks to achieve our goals, rather than design new architectures. Architecture details are provided in the \cref{sec:supp-neural-solver}.

Specifically, the joint feature extractor $\phi$ is built on an action recognition network~\cite{yan2018spatial}, which effectively extracts human motion information from sequential joint inputs. The input to $\phi$ is the concatenation of the initial and target joints in the shared human-centric coordinate system:
\begin{equation}
\label{eq:cat-joints}
    F = \phi(J_{init, init} \oplus J_{target, init}),
\end{equation}
where $F\in\mathbb{R}^D$ denotes the resulting joint feature vector in $D$ dimension and $\oplus$ indicates concatenation.

Then we further concatenate the initial SMPL parameters to supply joint features and input the combined representation into a lightweight MLP-based regressor $\psi$~\cite{wang2023refit}. 
Since inverse kinematics is inherently ill-posed, directly regressing the absolute target SMPL parameters often leads to suboptimal solutions. To mitigate this, we let the regressor $\psi$
to predict the residual pose parameters with respect to the initialization:
\begin{equation}
    \Delta\boldsymbol{\theta} = \psi(F\oplus\boldsymbol{\theta}_{init} \oplus \boldsymbol{\beta}).
\end{equation}
Practically, we represent the pose parameters using rotation matrices~\cite{goel2023humans,cai2023smpler,wang2023refit} for numerical continuity. The final prediction is obtained via rotation composition in $SO(3)$:
\begin{equation}
    \boldsymbol{\theta}_{pred} = \Delta\boldsymbol{\theta} \ \ \boldsymbol{\theta}_{init}.
\end{equation}

\subsubsection{Data Preparation and Inference Protocols}
\label{sec:implementation}
Up to this point, we have introduced how the Learnable SMPLify framework estimates the target pose parameter through a single-time regression.
Next, we present the construction of training data and the implementation of inference in our proposed framework.

\vspace{2mm}
\noindent
\textbf{Training.}
As a learning-based method, a key challenge is constructing meaningful paired data for training.
In Learnable SMPLify, this corresponds to selecting appropriate initialization-target pairs. 
The initialization must be close enough to the target to allow convergence in the ill-posed inverse kinematics problem, yet sufficiently distant to prevent the model from exploiting trivial solutions.
Therefore, by exploiting the continuity of human motion, we construct paired training data by sampling adjacent frames from motion sequences.
Given a motion sequence represented by SMPL parameters as $\{({\boldsymbol{\theta}_0, \boldsymbol{\beta}}),({\boldsymbol{\theta}_1, \boldsymbol{\beta}}), ...,({\boldsymbol{\theta}_{T-1}, \boldsymbol{\beta}})\}$, we uniformly sample training pairs at time index $t$ with a temporal offset $s$ as follows:
\begin{equation}
\label{eq:sampling}
    \boldsymbol{\theta}_{init} = \boldsymbol{\theta}_{t-s}, \ \ and \ \ \boldsymbol{\theta}_{target} = \boldsymbol{\theta}_t.
\end{equation}
To improve generalization and robustness, we vary the sampling interval $s$ during training within a range $[1, S]$, and apply data augmentation by swapping the initialization and target samples.

Based on these sample pairs, we supervise the model using constraints at three levels: pose parameters $\boldsymbol{\theta}$, 
regressed keypoints $J$, and human mesh vertices $V$. The corresponding losses are defined as:
\begin{equation}
    \mathcal{L}_{pose} = \arccos(\frac{Tr(\boldsymbol{\theta}_{target}^T\boldsymbol{\theta}_{pred}-1)}{2}),
\end{equation}
\begin{equation}
    \mathcal{L}_{kp} = ||J_{target} - J_{pred}||^2,
\end{equation}
\begin{equation}
    \mathcal{L}_{mesh} = ||V_{target} - V_{pred}||^2.
\end{equation}
Here, $\mathcal{L}_{pose}$ is a geodesic loss on $SO(3)$, while the others use $L2$ loss.

The total loss $\mathcal{L}$ is defined as:
\begin{equation}
    \mathcal{L} = \lambda_{pose}\mathcal{L}_{pose} + \lambda_{kp}\mathcal{L}_{kp} + \lambda_{mesh}\mathcal{L}_{mesh}.
\end{equation}
The weights $\lambda_{pose}, \lambda_{kp}, \lambda_{mesh}$ are hyperparameters that balance the influence of each term.

\vspace{2mm}
\noindent
\textbf{Inference.}
Learnable SMPLify supports two inference protocols:
\begin{itemize}
    \item \textit{Plug-in Post-processing}. Following the strategy used in SPIN~\cite{kolotouros2019spin}, any SMPL prediction can be treated as an initialization. The model then refines it by regressing residual SMPL parameters in a one-step forward.
    \item \textit{Sequential Inference}. For sequential data, given the first frame SMPL parameters, the model predicts the parameters for each subsequent frame by using the previous frame's output as initialization.
\end{itemize}
The sequential inference protocol reflects the core motivation behind our training data construction: modeling motion transitions from plausible initializations. This enables a robust application to sequential inputs by leveraging initializations that are naturally provided by temporal consistency.
Additionally, since the subject shape remains constant across frames, we simplify the pipeline by focusing on pose inverse kinematics, as formulated in \cref{eq:assumption}.
\section{Experiments}

\subsection{Datasets and Metrics}
We split the AMASS~\cite{mahmood2019amass} dataset into training and testing sets with a $7:3$ ratio, and use the training split to train Learnable SMPLify. For evaluation, we perform both in-domain testing on AMASS and cross-domain evaluation on 3DPW~\cite{von2018pw3d} and RICH~\cite{huang2022rich}.
Additionally, we further evaluate the plug-in post-processing ability on LucidAction~\cite{dong2024lucidaction}, a dataset of gymnastic scenes with challenging actions.
Dataset information is detailed in \cref{sec:supp-dataset}.

For evaluation metrics, we adopt Per-Vertex Error (PVE), which computes the mean $L_2$ distance between predicted and ground-truth SMPL vertices, to assess the human pose and shape estimation accuracy. We additionally report PA-PVE, which applies Procrustes Alignment before error computation to account for global misalignment.

\subsection{Implementation Details}
The temporal sampling range $S$ is set to $9$. The loss balance weights $\lambda_{pose}$, $\lambda_{kp}$ and $\lambda_{mesh}$ are set to $1.0$, $5.0$, and $1.0$, respectively.
We use the OpenPose~\cite{openpose} SMPL regressor to generate $25$ keypoints for the framework input.
The model is optimized using AdamW~\cite{loshchilov2017decoupled} with a batch size of 128 for 100 epochs. The learning rate is initialized at $10^{-4}$ and decayed via cosine annealing.

\subsection{Evaluation on Sampled Frames}
\label{sec:sampled-frames-exp}
\begin{table*}[t]
    \centering
 {
\captionof{table}{Evaluations on $s$-step prior frame initialization. PVEs are in $mm$. Runtime is in seconds. Lower is better for all metrics.}
\resizebox{\textwidth}{!}{
    \begin{tabular}{l|c|c|c|c|c|c|c|c|c|c|c|c|c|c}
    \toprule
    {\multirow{3}{*}{Method}} & \multicolumn{6}{c|}{AMASS} & \multicolumn{4}{c|}{3DPW} & \multicolumn{4}{c}{RICH} \\
    \cmidrule{2-15}
    {} & \multicolumn{3}{c|}{$s=1$} & \multicolumn{3}{c|}{$s=5$} & \multicolumn{2}{c|}{$s=1$} & \multicolumn{2}{c|}{$s=5$} & \multicolumn{2}{c|}{$s=1$} & \multicolumn{2}{c}{$s=5$}\\
    \cmidrule{2-15}
    {} & {PVE} & {PA-PVE} & {Runtime} & {PVE} & {PA-PVE} & {Runtime} & {PVE} & {PA-PVE} &{PVE} & {PA-PVE}  & {PVE} & {PA-PVE} &{PVE} & {PA-PVE}\\
    \midrule
    {Direct Copy} & {$21.36$} & {$9.72$} & {-} & {$98.72$} & {$44.02$} & {-} & {$24.51$} & {$9.38$} & {$106.85$} & {$40.84$} & {$31.08$} & {$13.04$} & {$95.82$} & {$42.14$}\\
    {SMPLify~\cite{smplify}} & {$18.85$} & {$18.50$} & {$11.73$} & {$19.03$} & {$18.56$} & {$12.52$} & {$17.21$} & {$17.08$} & {$18.31$} & {$18.06$} & {$20.75$} & {$21.19$} & {$21.05$} & {$21.45$} \\
    \cmidrule{1-15}
    {Ours} & {$\mathbf{3.23}$} & {$\mathbf{2.23}$} & {$\mathbf{0.06}$} & {$\mathbf{9.74}$} & {$\mathbf{7.00}$} & {$\mathbf{0.06}$} & {$\mathbf{4.35}$} & {$\mathbf{2.66}$} & {$\mathbf{13.09}$} & {$\mathbf{8.69}$} & {$\mathbf{12.57}$} & {$\mathbf{5.87}$} & {$\mathbf{20.92}$} & {$\mathbf{11.96}$}\\
    \bottomrule
  \end{tabular}
  \label{table:sample-frames}
}}
\end{table*}

\begin{table}[t]
    \centering
 {
\captionof{table}{Evaluations on sequential inference. PVEs are in $mm$. Lower is better for all metrics.}
\resizebox{\linewidth}{!}{
    \begin{tabular}{l|c|c|c|c|c|c}
    \toprule
    {\multirow{2}{*}{Method}} & \multicolumn{2}{c|}{AMASS} & \multicolumn{2}{c|}{3DPW} & \multicolumn{2}{c}{RICH} \\
    \cmidrule{2-7}
    {} & {PVE} & {PA-PVE} & {PVE} & {PA-PVE} & {PVE} & {PA-PVE} \\
    \midrule
    {Song~\etal~\cite{song2020human}} & {$21.61$} & {$13.29$} & {$104.56$} & {$51.29$} & {$115.13$} & {$58.11$} \\
    {KAMA~\cite{iqbal2021kama}} & {-} & {-} & {$47.40$} & {-} & {-} & {-} \\
    {SMPLify~\cite{smplify}} & {$28.00$} & {$26.40$} & {$26.83$} & {$24.71$} & {$38.95$} & {$31.64$} \\
    \midrule
    {Ours} & {$\mathbf{17.22}$} & {$\mathbf{15.72}$} & {$\mathbf{21.23}$} & {$\mathbf{19.34}$} & {$\mathbf{27.51}$} & {$\mathbf{24.02}$} \\
    \bottomrule
  \end{tabular}
  \label{table:sequential-infer}
}}
\end{table}

To evaluate inverse kinematics performance, we follow the data construction procedure outlined in~\cref{sec:implementation}, which samples both initialization and target poses in a sequence with a temporally offset defined by $s$. Specifically, the initialization pose is given by $\boldsymbol{\theta}_{t-s}$, and the target pose by $\boldsymbol{\theta}_{t}$.
As baselines, we include:
\begin{itemize}
    \item Direct Copy, which simply uses the initialization as the prediction to highlight discrepancies between the initial and target poses.
    \item SMPLify~\cite{smplify}, an optimization-based method that refines SMPL parameters from initialization.
\end{itemize}

In \cref{table:sample-frames}, we report performance on the in-domain AMASS dataset. Despite the close proximity between initialization and target poses, our method significantly outperforms the Direct Copy baseline, indicating that performance gains do not merely result from initialization similarity. Compared to SMPLify, our approach effectively utilizes initialization and captures fine-grained differences ($s=1$), leading to superior performance.

Moreover, our method achieves fast runtime by eliminating the costly iterative procedures. Specifically, it offers close to $200\times$ speed-up compared to SMPLify with only $6.1M$ trainable parameters, making it suitable for real-time applications. Unlike SMPLify, whose runtime increases from $11.73s$ at $s=1$ to $12.52s$ at $s=5$, our method’s inference time is invariant to initialization quality, maintaining both speed and accuracy consistently.

We further assess generalization ability, using the cross-domain 3DPW and RICH datasets. Even without being trained on these datasets, our method consistently outperforms SMPLify across varying initialization conditions.

\subsection{Evaluation on Sequential Inference}

Building on the frame-based evaluation, we further assess performance in the practical setting of sequential inference. As introduced in \cref{sec:implementation}, we initialize the sequence using the first frame and iteratively predict each subsequent frame based on the preceding prediction.
We include extra baselines for comparison:
\begin{itemize}
    \item Song~\etal~\cite{song2020human}, a learning-based approach that iteratively regresses poses from a canonical T-pose. We implement this under the same training scheme as our method. However, due to its network design, it cannot incorporate initialization directly into its predictions.
    \item KAMA~\cite{iqbal2021kama}, derives an analytical solution based on the human kinematic tree. Since KAMA is not open-sourced, we only include its reported performance as a reference.
\end{itemize}
As discussed in \cref{sec:human-ik}, our work focuses on solving the IK problem based on joint positions. Therefore, we do not include image-based regression methods as baselines for comparison, as they operate under distinct settings.

\begin{figure}
    \centering
    \includegraphics[width=\linewidth]{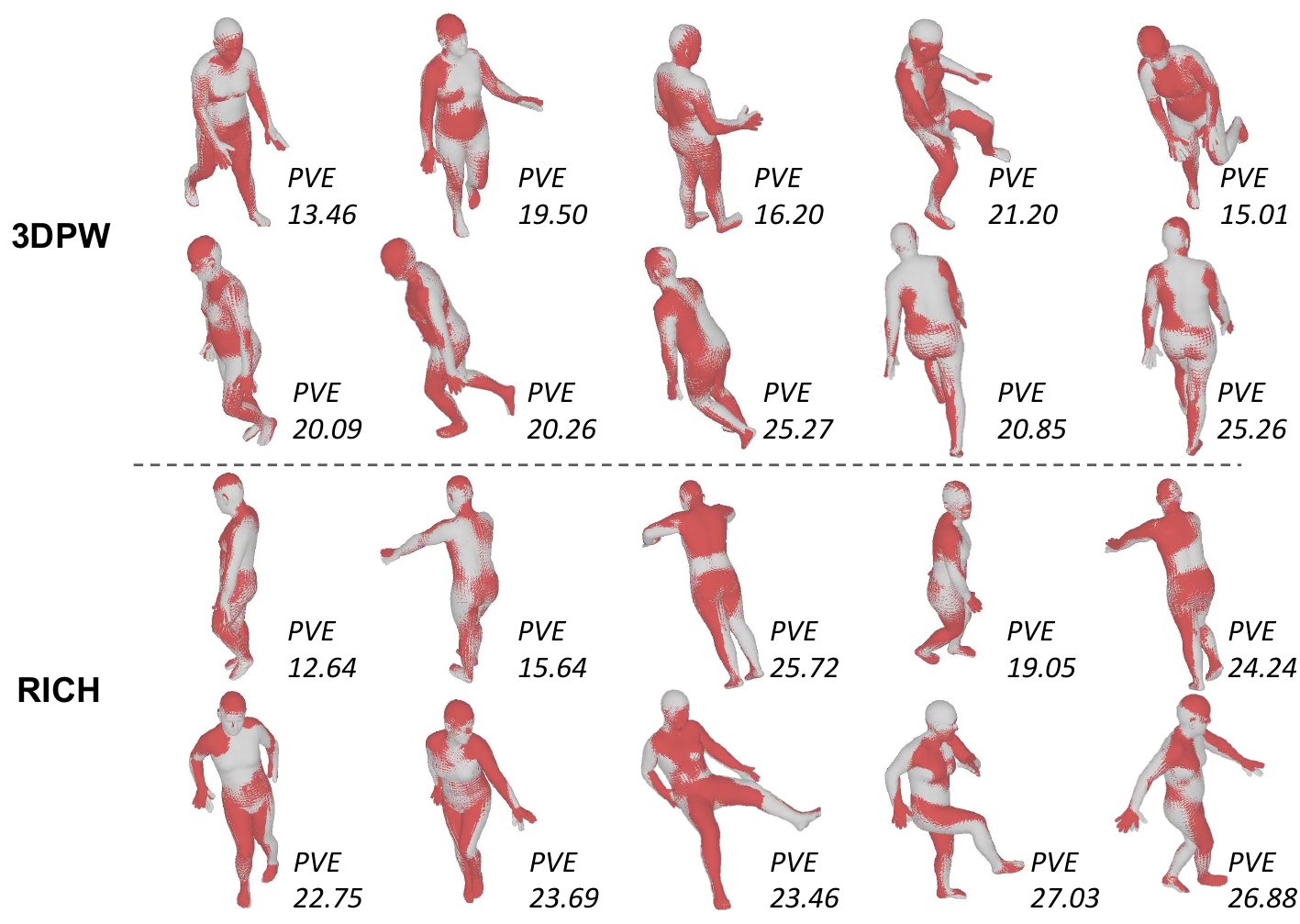}
    \caption{Qualitative evaluation on sequential inference. Predicted (white) and ground-truth meshes (red) are overlaid for visual comparison. Each row displays samples from a single sequence.}
    \label{fig:qualitative}
\end{figure}

\begin{figure*}
    \centering
    \includegraphics[width=\linewidth]{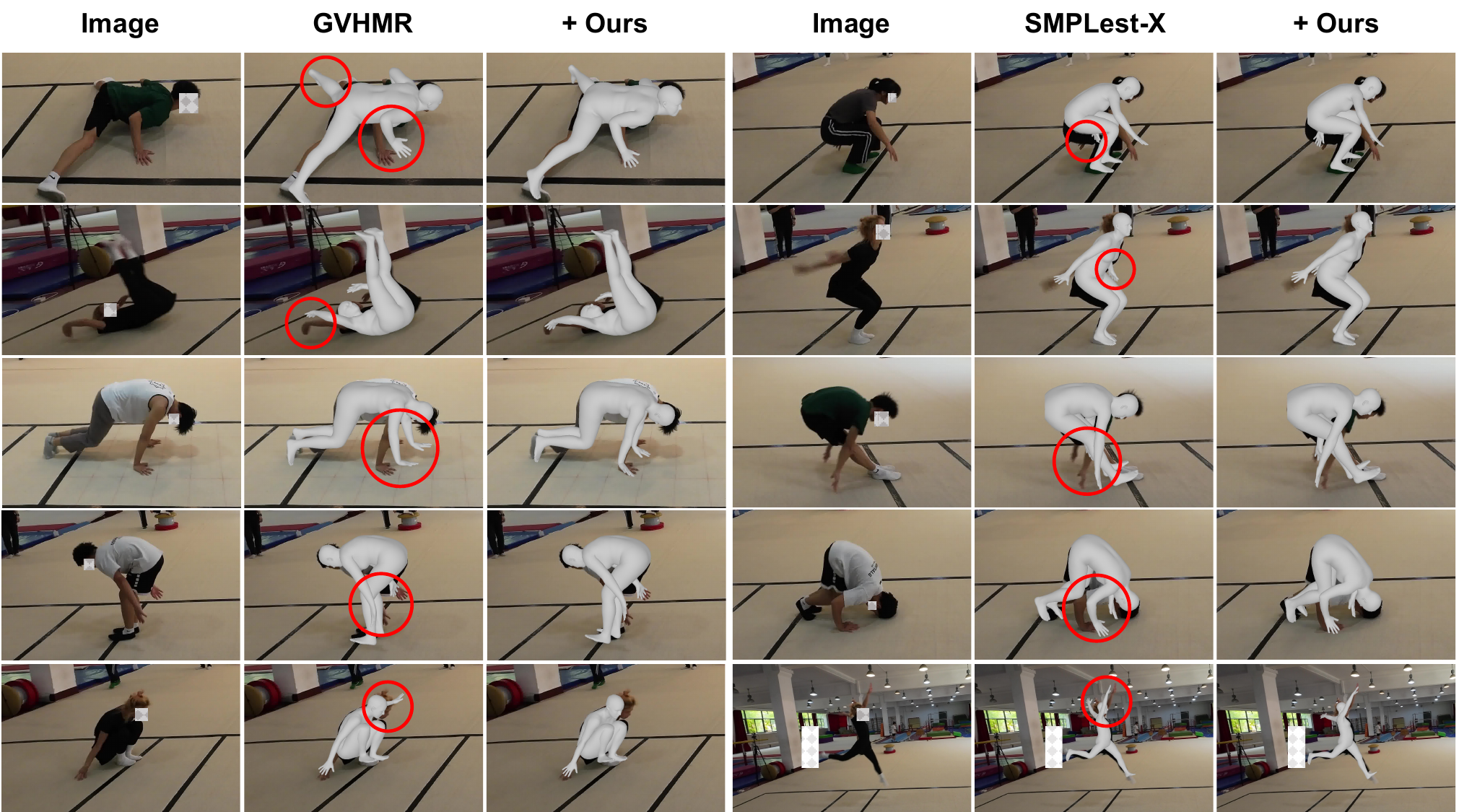}
    \caption{Qualitative results of plug-in post-processing. The left and right columns present the results based on GVHMR and SMPLest-X, respectively. Significant errors are marked with red circles. Faces are mosaicked for ethical considerations. Best viewed when zoomed in.}
    \label{fig:plug-in}
\end{figure*}

As shown in \cref{table:sequential-infer}, the proposed Learnable SMPLify consistently outperforms all baselines across evaluated datasets, 
achieving $17.22mm$, $21.23mm$, and $27.51mm$ PVE on AMASS, 3DPW, and RICH datasets. These results demonstrate robust consistency and generalization of our approach in a real application.
Notably, unlike Song~\etal, our method can effectively leverage initialization, yielding significant performance gains on previously unseen data in 3DPW and RICH datasets.
Similar to SMPLify, our method must address error accumulation arising from initialization based on preceding predictions during sequential inference. The superior performance demonstrates our model’s ability to produce plausible results while remaining robust to noise, even without relying on the iterative optimization used in SMPLify to correct predictions.

Qualitative results in \cref{fig:qualitative} further confirm the robustness of our method to obtain plausible predictions during sequential inference. As shown, the per-vertex error of approximately $20mm$ reported in \cref{table:sequential-infer} translates to only minor deviations at the scale of the human body with limited perceptual impact.
Combined with fast inference speed, the high visual accuracy of our method underscores its practicality for real-world motion capture applications.

\subsection{Evaluation on Plug-in Post-Processing}
As described in \cref{sec:implementation}, aside from sequential inference, the proposed Learnable SMPLify also supports plug-in post-processing. 
To evaluate its effectiveness in this setting, we adopt two state-of-the-art image-based human pose and shape estimation methods as baselines:
\begin{itemize}
    \item GVHMR~\cite{shen2024world}, which provides estimation by leveraging multiple 2D cues, including YOLO bonding boxes~\cite{yolov8}, 2D poses from ViTPose~\cite{xu2022vitpose}, and image features from HMR2.0~\cite{goel2023humans}.
    \item SMPLest-X~\cite{yin2025smplest}, which associates $10$ million training instances from diverse data sources for scaling.
\end{itemize}
We conduct experiments on the LucidAction~\cite{dong2024lucidaction} dataset, which is not included in the training sets of most existing human pose and shape estimation methods. This provides a realistic evaluation setting involving complex, cross-domain motion sequences with challenging poses.

As shown in~\cref{fig:plug-in}, the baseline methods exhibit noticeable errors,
especially in end-effectors such as hands and feet, largely due to error accumulation propagated through the kinematic tree.
Our method takes the baseline predictions as initialization and effectively refines the results. Notably, its performance across both baselines unveils its model-agnostic nature, enabled by generalization from temporal sequence training.
For the LucidAction dataset, joint labels are obtained using a 2D estimator~\cite{jiang2023rtmpose} combined with multi-view triangulation, which introduces a certain level of noise. Our approach’s robustness to such noise makes our approach a practical and lightweight plug-in post-processing solution that seamlessly leverages the rapid progress in image-based human pose and shape estimation.

\subsection{Ablation Study}

\begin{table*}[t]
    \centering
 {
\captionof{table}{Ablation study on framework components with $s$-step prior frame initialization. PVEs are in $mm$. Lower is better for all metrics.}
\resizebox{\linewidth}{!}{
    \begin{tabular}{c|l|c|c|c|c|c|c|c|c}
    \toprule
    {\multirow{3}{*}{Exp.}} & {\multirow{3}{*}{Setting}} & \multicolumn{4}{c|}{AMASS} & \multicolumn{4}{c}{3DPW} \\
    \cmidrule{3-10}
    {} & {} & \multicolumn{2}{c|}{$s=1$} & \multicolumn{2}{c|}{$s=5$} & \multicolumn{2}{c|}{$s=1$} & \multicolumn{2}{c}{$s=5$}\\
    \cmidrule{3-10}
    {} & {} & {PVE} & {PA-PVE} &{PVE} & {PA-PVE}  & {PVE} & {PA-PVE} &{PVE} & {PA-PVE}\\
    \midrule
    {1} & {\textit{w/o} Human-centric Normalization} & {$10.28$} & {$5.41$} & {$22.47$} & {$13.08$} & {$59.63$} & {$13.93$} & {$115.75$} & {$41.17$} \\
    {2} & {Direct Prediction} & {$35.95$} & {$22.26$} & {$35.95$} & {$22.26$} & {$276.28$} & {$102.48$} & {$276.28$} & {$102.48$} \\
    {-} & {Learnable SMPLify} & {$\mathbf{3.28}$} & {$\mathbf{2.28}$} & {$\mathbf{9.80}$} & {$\mathbf{7.06}$} & {$\mathbf{4.40}$} & {$\mathbf{2.69}$} & {$\mathbf{13.23}$} & {$\mathbf{8.80}$} \\
    \bottomrule
  \end{tabular}
  \label{table:abl-component}
}}
\end{table*}

\begin{table}[t]
    \centering
 {
\captionof{table}{Ablation study on temporal sampling range $S$ with $s$-step prior frame initialization. PVEs are in $mm$. Lower is better for all metrics.}
\resizebox{\linewidth}{!}{
    \begin{tabular}{l|c|c|c|c|c|c|c|c}
    \toprule
    {\multirow{3}{*}{Setting}} & \multicolumn{4}{c|}{AMASS} & \multicolumn{4}{c}{3DPW} \\
    \cmidrule{2-9}
    {} & \multicolumn{2}{c|}{$s=1$} & \multicolumn{2}{c|}{$s=5$} & \multicolumn{2}{c|}{$s=1$} & \multicolumn{2}{c}{$s=5$}\\
    \cmidrule{2-9}
    {} & {PVE} & {PA-PVE} &{PVE} & {PA-PVE}  & {PVE} & {PA-PVE} &{PVE} & {PA-PVE}\\
    \midrule
    {$S=3$} & {$3.30$} & {$2.17$} & {$15.04$} & {$8.96$} & {$4.56$} & {$2.72$} & {$19.41$} & {$11.13$} \\
    {$S=5$} & {$3.49$} & {$2.35$} & {$12.70$} & {$8.52$} & {$4.59$} & {$2.80$} & {$16.94$} & {$10.43$} \\
    {$S=7$} & {$3.81$} & {$2.41$} & {$12.76$} & {$8.20$} & {$4.99$} & {$2.88$} & {$16.54$} & {$10.00$} \\
    {$S=9$} & {$\mathbf{3.28}$} & {$\mathbf{2.28}$} & {$\mathbf{9.80}$} & {$\mathbf{7.06}$} & {$\mathbf{4.40}$} & {$\mathbf{2.69}$} & {$\mathbf{13.23}$} & {$\mathbf{8.80}$} \\
    {$S=11$} & {$3.91$} & {$2.57$} & {$12.15$} & {$8.12$} & {$5.24$} & {$3.04$} & {$16.41$} & {$9.96$} \\
    \bottomrule
  \end{tabular}
  \label{table:abl-stride}
}}
\end{table}

To evaluate the effectiveness of the proposed strategies in Learnable SMPLify, we conduct ablation studies in this section. For efficient evaluation, all models are trained using 50\% of the AMASS training split. We evaluate them on the AMASS and 3DPW test splits to represent in-domain and cross-domain scenarios, respectively.
We adopt the experiment setting on sampled frames described in~\cref{sec:sampled-frames-exp} to assess the fundamental inverse kinematics capability under varying initialization conditions.

\subsubsection{Effectiveness of the Framework}
As shown in \cref{table:abl-component}, we focus our analysis on two key components of the proposed framework: Human-centric Normalization, which aims to improve generalization across domains, and residual learning in the Neural Solver, which is designed to effectively leverage initialization for further generalization improvement in inverse kinematics. 

In Exp.1, we replace the proposed Human-centric Normalization with a simple baseline that applies location and scale normalization to both the initialization and target joints. While this baseline achieves reasonable performance on the AMASS dataset, its effectiveness drops significantly on the 3DPW dataset, with the PVE increasing to $115.75 mm$ under the $s=5$ setting. This degradation highlights the domain shift that remains unresolved by naive normalization. In contrast, Learnable SMPLify generalizes well to 3DPW, demonstrating that the proposed Human-centric Normalization effectively mitigates cross-domain discrepancies.

In Exp.2, we evaluate the Direct Prediction baseline by using a fixed T-pose as initialization, replacing the $s$-step prior frame. As a result, its performance remains unchanged across different values of $s$. However, this direct prediction strategy results in inferior performance, particularly on the unseen 3DPW dataset, where the PVE increases to $276 mm$. This degradation is primarily due to the ambiguity introduced when estimating target poses from a fixed T-pose, which reflects the ill-posed nature of human pose inverse kinematics.
In comparison, Learnable SMPLify eases this issue by utilizing temporally consistent initialization, highlighting the effectiveness of the residual learning design in the proposed Neural Solver.

\subsubsection{Impact of Temporal Sampling Range}
As shown in \cref{table:abl-stride}, we analyze the impact of temporal sample range $[1, S]$, which is implemented in \cref{eq:sampling} to introduce diversity and robustness into the training process.
By varying the temporal sampling range $S \in \{3, 5, 7, 9, 11\}$, we observe that Learnable SMPLify achieves the best overall performance on both the AMASS and 3DPW datasets when $S=9$. For a smaller sampling range, the model learns to handle small discrepancies between the initialization and target poses effectively. For example, the $S=3$ setting performs well under the $s=1$ condition but struggles when faced with larger temporal gaps such as $s=5$. In contrast, a larger sampling range leads to the opposite effect. The $S=9$ configuration offers a balanced trade-off, providing an optimal point to allow the model to generalize across both short- and long-term transitions.
\section{Conclusion}
In this paper, we propose Learnable SMPLify, a neural solution for optimization-free human pose inverse kinematics. By replacing the iterative optimization in traditional SMPLify with a neural solver, our approach significantly accelerates inference speed while maintaining high accuracy. To enhance generalization ability, we introduce a human-centric normalization strategy on both initial and target joints. Training pairs are constructed by sampling from temporal sequences, enabling the model to tackle the ill-posed inverse kinematics via residual learning.
Benefited from these designs, Learnable SMPLify produces consistent and plausible predictions in sequential inference and exhibits a model-agnostic nature in plug-in post-processing for refining methods. 
Extensive experiments across multiple benchmarks validate the effectiveness of Learnable SMPLify as a practical and simple baseline for human pose inverse kinematics.

{
    \small
    \bibliographystyle{ieeenat_fullname}
    \bibliography{main}
}
\clearpage
\setcounter{page}{1}
\maketitlesupplementary
\appendix

\section{Neural Solver Architecture}
\label{sec:supp-neural-solver}
The neural solver in Learnable SMPLify comprises a feature extractor and a regressor.

The feature extractor $\phi$ builds on the classic skeleton-based action recognition method ST-GCN~\cite{yan2018spatial}. It processes input joint sequences $J_{seq} \in \mathbb{R}^{T\times N\times D}$, where the concatenated initial and target joints in \cref{eq:cat-joints} form a $T=2, N=25, D=3$ sequence, facilitating straightforward implementation. The extractor consists of $10$ blocks of graph convolution and temporal convolution layers, each with residual connections. It raises features $D$ from $3$ to $256$, then applies pooling over the temporal and joint dimensions to yield a compact joint-level representation $F\in \mathbb{R}^{256}$. Next, we concatenate the initial pose and beta parameters, resulting in an enriched feature vector of dimension $F\in \mathbb{R}^{335}$. For regressor $\psi$, we follow the design of ReFit~\cite{wang2023refit}. A two-layer MLP mapping the refined feature into $F\in\mathbb{R}^{M\cdot 256}$, where $M$ denotes the number of SMPL pose parameters. Finally, $M$ separate linear layers independently predict each pose parameter in a 6D rotation representation, corresponding to the first two columns of a rotation matrix.

\section{Experiment}
\subsection{Dataset Details}
\label{sec:supp-dataset}

\noindent
\textbf{AMASS.} AMASS associates various optical marker-based motion capture datasets to construct more than $40$ hours motion data, forming a rich foundation for our neural inverse kinematics solver. We split the dataset into train and test sets at the granularity of sequence with a $7:3$ ratio. To avoid repetitive initial poses, the first and last $10\%$ of data in each sequence are removed. All sequences are sampled at $30$ fps to ensure that our training strategy captures similar discrepancies between initialization and target. During training and testing, we sample initialization and target pairs at intervals of $10$ across all sequences. Finally, the data process yields $298,496$ initialization-target pairs for training and $127,926$ pairs for testing.

\noindent
\textbf{3DPW.} 3DPW collects human motion data in the wild using IMUs and a moving camera. It provides SMPL annotations in high quality that are widely used for evaluation in human pose and shape estimation. We only use the official test split to evaluate our model. It yields $35,145$ initialization-target pairs across $37$ sequences for testing.

\noindent
\textbf{RICH.} RICH contains motion data of real human-scene interactions, both indoors and outdoors, captured using a multi-view motion capture system with 3D scans. The dataset contains high-quality SMPL-X annotations, which we convert to SMPL format for evaluation. We use $298,638$ initialization-target pairs in $52$ sequences of the official test split to evaluate our model.

\noindent
\textbf{LucidAction.} LucidAction provides $259$ types of actions from $8$ diverse gymnastics events across $4$ distinct curriculum level, from beginner to expert difficulty. It provides 3D human pose annotations captured in multi-view optical motion capture systems.
We adopt $21$ types of actions from the second difficult level in men's/women's floor exercise events for demonstration.

\subsection{Experiment Details}
\noindent
\textbf{Sequential Inference.} As we focus on pose inverse kinematics and with the assumption that the subject's shape is invariant, we use the ground truth shape parameter for all baselines during evaluation to ensure a fair comparison. 
All initial and target joints are taken from the ground truth. This ensures an accurate evaluation of the algorithm’s ability to fit the SMPL pose to target data, while avoiding the influence of estimation noise. Later, the Plug-in Post-Processing step demonstrates a practical scenario where estimated joints are applied.
For SMPLify, we employ the implementation from ZJU-Mocap~\cite{easymocap}, which supports optimizing SMPL parameters on 3D joints. We initialize both SMPLify and our Learnable SMPLify with the ground-truth pose of the first frame, and apply frame-by-frame inference for the full sequence. For Song~\etal's we apply the 3D joints as input and train their model using the same AMASS dataset split as ours to ensure fairness in evaluation. For KAMA, we select the reported performance under the setting that uses ground truth joints to ensure alignment with our evaluation protocol.

\noindent
\textbf{Plug-in Post-Processing.} Although GVHMR is a world-aligned method, we observe scale estimation errors when comparing its outputs with the 3D joints provided in LucidAction. To address this, we use the results in the estimated camera coordinate system and transform them into the world coordinate system using the calibrated camera parameters. Alternatively, the same alignment can be achieved by transforming the ground-truth joints into the estimated camera coordinate system.

For SMPLest-X, we use the publicly released large-scale model for experiments. Similarly, we convert its predictions into SMPL format in the world coordinate system for convenience.

\subsection{Iterative Sequential Inference}

\begin{table}[t]
\centering
 {
\captionof{table}{Evaluations on iterative sequential inference. PVEs are in $mm$. Lower is better for all metrics.}
\resizebox{\linewidth}{!}{
    \begin{tabular}{l|c|c|c|c|c|c}
    \toprule
    {\multirow{2}{*}{Config.}} & \multicolumn{2}{c|}{AMASS} & \multicolumn{2}{c|}{3DPW} & \multicolumn{2}{c}{RICH} \\
    \cmidrule{2-7}
    {} & {PVE} & {PA-PVE} & {PVE} & {PA-PVE} & {PVE} & {PA-PVE} \\
    \midrule
    {$N=1$} & {${17.22}$} & {${15.72}$} & {${21.23}$} & {${19.34}$} & {${27.51}$} & {${24.02}$} \\
    {$N=3$} & {$16.25$} & {$14.97$} & {$\mathbf{19.92}$} & {$\mathbf{18.24}$} & {$25.72$} & {$22.99$} \\
    {$N=5$} & {$\mathbf{15.80}$} & {$\mathbf{14.56}$} & {$20.03$} & {$18.28$} & {$\mathbf{25.52}$} & {$\mathbf{22.93}$} \\
    \bottomrule
  \end{tabular}
  \label{table:iter}
}}
\end{table}

Possessing a single forward characteristic, the proposed Learnable SMPLify can also be applied iteratively. In \cref{table:iter}, we evaluate the performance on sequential inference. For each frame, we perform $N$ iterations, where the result from the $k-1$th iteration is used to initialize the  $k$-th iteration. The initialization for the $0$-th iteration of the current frame is taken from the final prediction of the previous frame.
We present the best results between iterative inference and non-iterative inference for each configuration. The comparison of $N=1$ and $N=3$ settings demonstrates that applying our method iteratively further improves the performance. The comparison between $N=3$ and $N=5$ indicates convergence, as increasing the number of iterations yields no significant and consistent gains.

\end{document}